\documentclass[acmtog,screen]{acmart}
\AtBeginDocument{%
  }

\newif\ifanonymized
\anonymizedfalse

\newif\ifconfidential
\confidentialfalse

\newif\ifauthorversion
\authorversionfalse

\newif\ifredlined
\redlinedfalse

\usepackage{subcaption} 
\usepackage{multirow}
\usepackage{stfloats}
\usepackage{balance}
\usepackage{algorithm}
\usepackage{algpseudocode}
\usepackage{amsmath}
\usepackage{xcolor}
\usepackage{xspace}
\usepackage{soul}
\usepackage{makecell}
\usepackage{comment}
\usepackage[utf8]{inputenc}

\usepackage{mathtools}

\definecolor{darkgreen}{rgb}{0,0.5,0}
\definecolor{orange}{rgb}{1,0.5,0}
\definecolor{purple}{rgb}{0.5,0.2,0.8}
\definecolor{darkred}{rgb}{0.5,0,0}
\definecolor{Yellow}{rgb}{1,1, 0.6}
\definecolor{Orange}{rgb}{1,0.8, 0.6}
\definecolor{lightgray}{gray}{0.96}

\newcommand\nth{\textsuperscript{th}\xspace} 

\ifredlined
    \newcommand{\redline}[1]{\textcolor{red}{\st{#1}}}
    \newcommand{\blueline}[1]{\textcolor{blue}{#1}}
    \sethlcolor{Yellow}
    
\else
    \newcommand{\redline}[1]{}
    \newcommand{\blueline}[1]{#1}
    \sethlcolor{Yellow}
    
\fi

\copyrightyear{2025}
\acmYear{2025}
\setcopyright{rightsretained} 
\acmConference[SIGGRAPH Conference Papers '25]{Special Interest Group on Computer Graphics and Interactive Techniques Conference Conference Papers }{August 10--14, 2025}{Vancouver, BC, Canada}
\acmBooktitle{Special Interest Group on Computer Graphics and Interactive Techniques Conference Conference Papers (SIGGRAPH Conference Papers '25), August 10--14, 2025, Vancouver, BC, Canada}
\acmDOI{10.1145/3721238.3730763}
\acmISBN{979-8-4007-1540-2/2025/08}

\acmConference[SIGGRAPH Conference Papers '25]{Special Interest Group on Computer Graphics and Interactive Techniques Conference Conference Papers}{August 10--14, 2025}{Vancouver, BC, Canada}
\acmBooktitle{Special Interest Group on Computer Graphics and Interactive Techniques Conference Conference Papers (SIGGRAPH Conference Papers '25), August 10--14, 2025, Vancouver, BC, Canada}

\begin{document}

\title{Fast Isotropic Median Filtering}

\author{Ben Weiss}
\orcid{0009-0001-0395-0242}
\affiliation{
    \institution{Google Research}
    \streetaddress{1600 Amphitheatre Parkway}
    \city{Mountain View}
    \state{CA}
    \country{USA}
}
\authorsaddresses{
    Authors’ address:
    \href{https://www.benweiss.com/fast-isotropic-median-filtering/index.html}{Ben Weiss},
    Google Inc., 1600 Amphitheatre Parkway, Mountain View, CA, 94043, USA
}
\renewcommand{\shortauthors}{Ben Weiss}

\begin{teaserfigure}
  \centering
  \includegraphics[width=\textwidth]{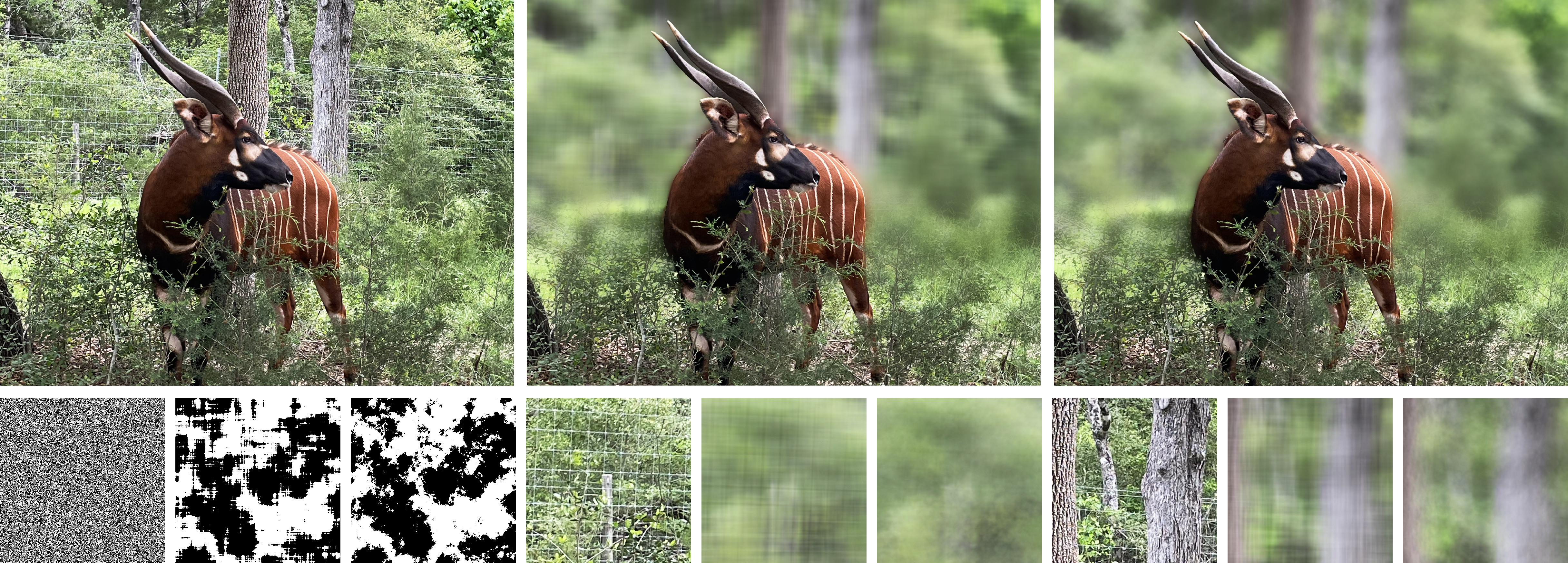}
  \caption{Top row: Original photo, median-filtered with a radius-48 square kernel (with hand-applied matte), and with an equivalent-area circular kernel. Bottom row: square vs. circular filtered image quality for binary noise, as well as for two swatches from the image, unsharp-masked 200\% to emphasize the high-frequency artifacts. Across a wide range of parameters, our circular median filter is dramatically faster and higher quality than the state of the art.}
  \label{fig:teaser}
  \Description[Side-by-side photo comparison illustrating our method's quality improvement.]{Side-by-side illustration of an original photo of a bongo (African antelope) standing in front of some trees; the same photo median-filtered with square kernel; and then median-filtered with circular kernel, illustrating the image quality improvement obtained by using a circular kernel instead of square. Across the bottom of the figure are smaller swatches illustrating square-versus-circular-kernel image quality differences: first for pure binary noise, and then for two zoomed-in representative samples from the photo.}
\end{teaserfigure}

\begin{abstract}
Median filtering is a cornerstone of computational image processing. It provides an effective means of image smoothing, with minimal blurring or softening of edges, invariance to monotonic transformations such as gamma adjustment, and robustness to noise and outliers. However, known algorithms have all suffered from practical limitations: the bit depth of the image data, the size of the filter kernel, or the kernel shape itself. Square-kernel implementations tend to produce streaky cross-hatching artifacts, and nearly all known efficient algorithms are in practice limited to square kernels. We present for the first time a method that overcomes all of these limitations. Our method operates efficiently on arbitrary bit-depth data, arbitrary kernel sizes, and arbitrary convex kernel shapes, including circular shapes.
\end{abstract}



\maketitle

\section{Introduction}
\label{sec:introduction}

The median filter \cite{tukey1974nonlinear} is a foundational element of classical signal processing. Extended to two dimensions \cite{10.1109/TASSP.1979.1163188}, it allows removal of image noise and smoothing of edges, without the softening that is characteristic of e.g., Gaussian blur. Historically it has been fairly inefficient and artifact-prone; the nonlinear nature of the filter has resisted optimization attempts through vectorization techniques, and the ubiquitous square-kernel implementation produces distinct cross-hatching visual artifacts that reduce the filter's usefulness for quality-sensitive applications.

In this paper, we present an algorithm that performs efficient median filtering using a circular kernel, substantially improving image quality while attaining state-of-the-art performance for medium and large kernel sizes. Our primary contributions are as follows:
\begin{itemize}
    \item Extension of the 1D compound histogram technique from Weiss \shortcite{10.1145/1141911.1141918} to a more general and powerful 2D "omnigram" data structure, from which histogram elements of any ordinal image region of any shape can be obtained in constant time.
    \item A rank-order-filtering algorithm capable of processing circular kernel shapes, eliminating the visual artifacts characteristic of square-kernel filtering, while achieving state-of-the-art performance, even compared to the best existing square-kernel implementations.
\end{itemize}
After reviewing related work in section \ref{sec:related_work}, we give an overview of our method in section \ref{sec:system_overview}. We then describe our implementation in detail in section \ref{sec:implementation} and provide results and comparative studies in section \ref{sec:results}. Code and data for this paper are at:

\href{https://github.com/google/fast-isotropic-median-filter}{https://github.com/google/fast-isotropic-median-filter}.
\section{Related Work}
\label{sec:related_work}

\begin{figure*}
  \centering
  \includegraphics[width=\textwidth]{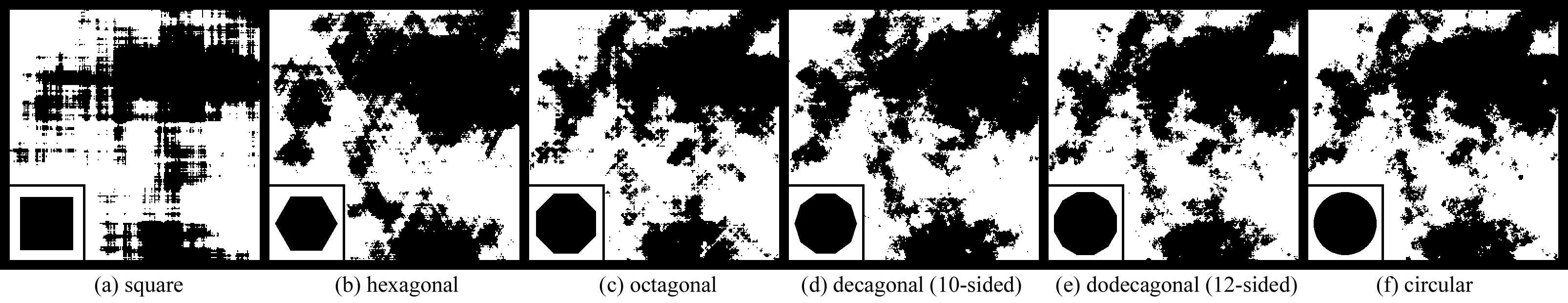}
  \caption{Median-filtering binary noise with increasingly isotropic kernel shapes. By observing the cross-hatching artifacts introduced by filtering noise with low-order polygonal kernels, the importance of using near-isotropic kernels (ideally 12-sided or circular) to obtain optimal image quality is strikingly evident.}
  \label{fig:median_polygon_noise_collage}
  \Description[Comparison of median filter artifacts for various polygonal and circular kernels.]{Six images are shown side-by-side, illustrating the results of median-filtering pure binary noise with 4-, 6-, 8-, 10-, 12-sided polygonal kernels and a circular kernel, respectively. The visual cross-hatching artifacts are very strong in the square-kernel case and become less and less noticeable as the polygons gain more sides, until the artifacts are barely perceptible in the 12-sided-kernel case, and imperceptible in the circular-kernel case.}
\end{figure*}

\noindent
{\centering
\begin{tabular}{|c|c|c|}
\hline
\textbf{Approach} & \textbf{Runtime} & \textbf{Implementations} \\
\hline
\makecell[c]{Sorting Networks} & \(\llap{$\sim$\hspace{2pt}} O(r^2)\) & \makecell[c]{McGuire \shortcite{10.2312/HPG/HPG08/059-066} \\ Adams \shortcite{10.1145/3450626.3459773}} \\
\hline
\makecell[c]{Sliding-Window \\ Histograms} & \makecell[c]{\(O(r)\), \\ \(O(\log^k(r))\),  \\ \(O(1)\),  \\ \(\llap{$\sim$\hspace{2pt}} O(r)\)} & \makecell[c]{Huang et al. \shortcite{10.1109/TASSP.1979.1163188} \\ Weiss \shortcite{10.1145/1141911.1141918} \\ Perreault and Hébert \shortcite{10.1109/TIP.2007.902329} \\ Our method} \\
\hline
\makecell[c]{Integral Queries} & \(O(1)\) & \makecell[c]{Moroto and Umetani \shortcite{10.1145/3550454.3555512}} \\
\hline
\makecell[c]{Isotropic \\ Approximations} & \(O(1)\) & \makecell[c]{Kass and Solomon \shortcite{10.1145/1778765.1778837} \\ Yang et al. \shortcite{10.1007/s11263-014-0764-y}} \\
\hline
\end{tabular}
\par}
\label{tab:median_filtering_techniques}

\subsection{Sorting Networks}

Sorting networks were introduced by Batcher \shortcite{10.1145/1468075.1468121}, where inputs are sorted directly using a fixed sequence of min-max swaps to obtain the median. McGuire \shortcite{10.2312/HPG/HPG08/059-066} pioneered this approach for 3x3 and 5x5 kernels, and Adams \shortcite{10.1145/3450626.3459773} extended it to larger kernels using separable sorting networks. For square kernels of fairly small size, this is the most efficient known technique. However, it becomes impractically slow for large kernels.

\subsection{Sliding-Window Histograms}

Other methods involve explicit histograms that are maintained and updated using sliding windows. Originated by Huang et al. \shortcite{10.1109/TASSP.1979.1163188}, this provides a canonical baseline \(O(r)\) algorithm for 8-bit images. In our previous work ( \cite{10.1145/1141911.1141918}) we had improved on the sliding-window technique for 8-bit images by constructing hierarchical trees of histograms, improving the runtime from \(O(r)\) to \(O(log(r))\), and also extended the method to higher bit-depths in \(O(log^2(r))\) time. Perreault and Hébert \shortcite{10.1109/TIP.2007.902329} employed a sliding-window technique over single-column histograms, yielding a constant-time median filtering algorithm for 8-bit data.   

\subsection{Integral Queries}

Porikli \shortcite{10.1109/CVPR.2005.188} first recognized that small (e.g. 8-bit) histograms could be spatially integrated, yielding summed-area-table-like structures from which histogram components of any rectangular region could be extracted in constant time. More recently, Moroto and Umetani \shortcite{10.1145/3550454.3555512} extended the 1D wavelet matrix technique of Claude et al. \shortcite{10.1016/j.is.2014.06.002} to process arbitrary-depth 2D images, allowing efficient and arbitrary 2D rectangular rank-order queries and median filtering in constant time.

\subsection{Isotropic Approximations}

The characteristic cross-hatching artifacts [Fig. ~\ref{fig:median_polygon_noise_collage}(a)] of square-kernel median filters have long been recognized, but the algorithmic costs of avoiding these artifacts have been difficult to overcome, particularly for exact filters. Thus, most current implementations (e.g. OpenCV, Adobe Photoshop) support only square-kernel median filtering. Mishra et al. \shortcite{Mishra2020ACircular} described the qualitative advantages of using an approximately circular kernel to median-filter medical MRI images, although they were limited to very small kernels.

Perreault and Hébert \shortcite{10.1109/TIP.2007.902329} and Moroto and Umetani \shortcite{10.1145/3550454.3555512} described extensions to their methods that support higher-order polygonal kernels, but at a linearly increasing performance cost (and working set overhead) in the number of sides.

Kass and Solomon \shortcite{10.1145/1778765.1778837} developed a constant-time approximate method for isotropic median filtering, by low-pass-filtering histogram data into a continuous representation, then applying a spatial Gaussian filter to achieve an isotropic weighted kernel shape. Yang et al. \shortcite{10.1007/s11263-014-0764-y} proposed a similar method involving coarse-granularity cost volumes, interpolated to find approximate medians. Both methods were targeted at 8-bit images; it is unclear how well they might practically scale to e.g. HDR domains.

Finally, in our earlier paper \cite{10.1145/1141911.1141918}, we mentioned as a topic of future research that our 1D "compound histogram" technique might be extended to accommodate non-rectangular kernel shapes, such as circles, but did not offer a practical implementation. In this paper we show that this conceptual approach can indeed work, and that it can be implemented efficiently and powerfully, with state-of-the-art performance and quality and a small memory footprint, on both GPU and CPU architectures.
\section{Method overview}
\label{sec:system_overview}

\subsection{Core concepts: Histograms, Pivots, and Counts}

In a histogram, the median value is found at the smallest index for which the cumulative histogram up to that point reaches the midpoint rank; e.g. for a 7x7 median filter, the 25th of 49 values. This can be found by scanning the histogram from its endpoints, but when the median must be extracted from two similar histograms (as for overlapping image windows), as observed by Huang et al., it is more efficient to choose a "pivot" index near the median value of the first region, for which the "count" (cumulative number of values strictly below the pivot) is known, then track how the "count" changes as the histogram is updated for the second region. The pivot from the first region is then used as the starting index to scan the updated histogram to find the median value for the second region.

\begin{figure}
  \centering
  \includegraphics[width=0.7\columnwidth]{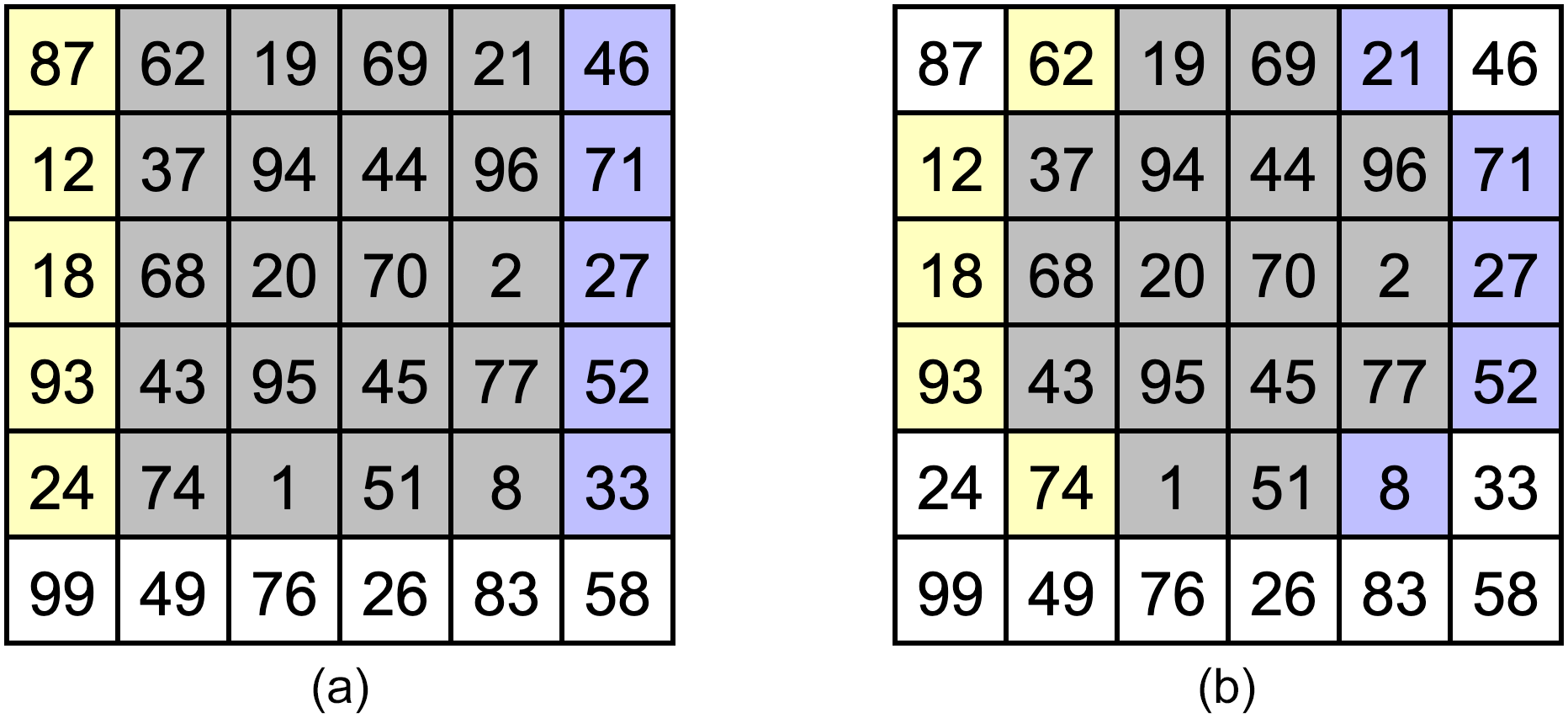}
  \caption{Sliding-window method of Huang et al. \shortcite{10.1109/TASSP.1979.1163188}. (a) As the top left 5x5 window slides one pixel to the right, blue-shaded pixels are added to a running histogram, and yellow-shaded pixels removed. (b) The method as adapted to a 21-tap circular kernel.}
  \label{fig:huang_circular}
  \Description[Illustration of sliding a small filter kernel one pixel to the right, for square and circular kernels.]{A 6x6 grid of numbers is shown on both the left (a), and right (b). In (a) two overlapping 5x5 squares are visualized within the 6x6 grid. Pixels that are only in the first 5x5 square (a 1x5 column in the top left) is tinted yellow; pixels common to both 5x5 squares (a 4x5 block in the top middle) are tinted gray, and pixels that are only in the second 5x5 square (a 1x5 column in the top right) are tinted blue. In part (b) on the right, the same idea is shown, but for an approximately circular kernel inset in a 5x5 square, consisting of 21 pixels (the 5x5 square excluding the corner pixels).}
\end{figure}

Huang's method (Fig. \ref{fig:huang_circular}) is linear-time in the filter radius, and works by sliding a square window across the image, following this pivot/count principle. It has the advantage of simplicity, and can be straightforwardly adapted to process non-square kernel shapes, such as circular shapes\footnote{We are not sure whether Huang et al. were aware of this.}. Its main drawback is that its primary bottleneck, the histogram-updating step, is fundamentally scalar and memory-bound. A second drawback is that its memory requirements grow exponentially with bit-depth, and also linearly with concurrency; each thread must maintain its own separate mutable histogram.

\subsection{The Ordinal Transform and Compound Histogram}

\begin{figure}
  \centering
  \includegraphics[width=0.6\columnwidth]{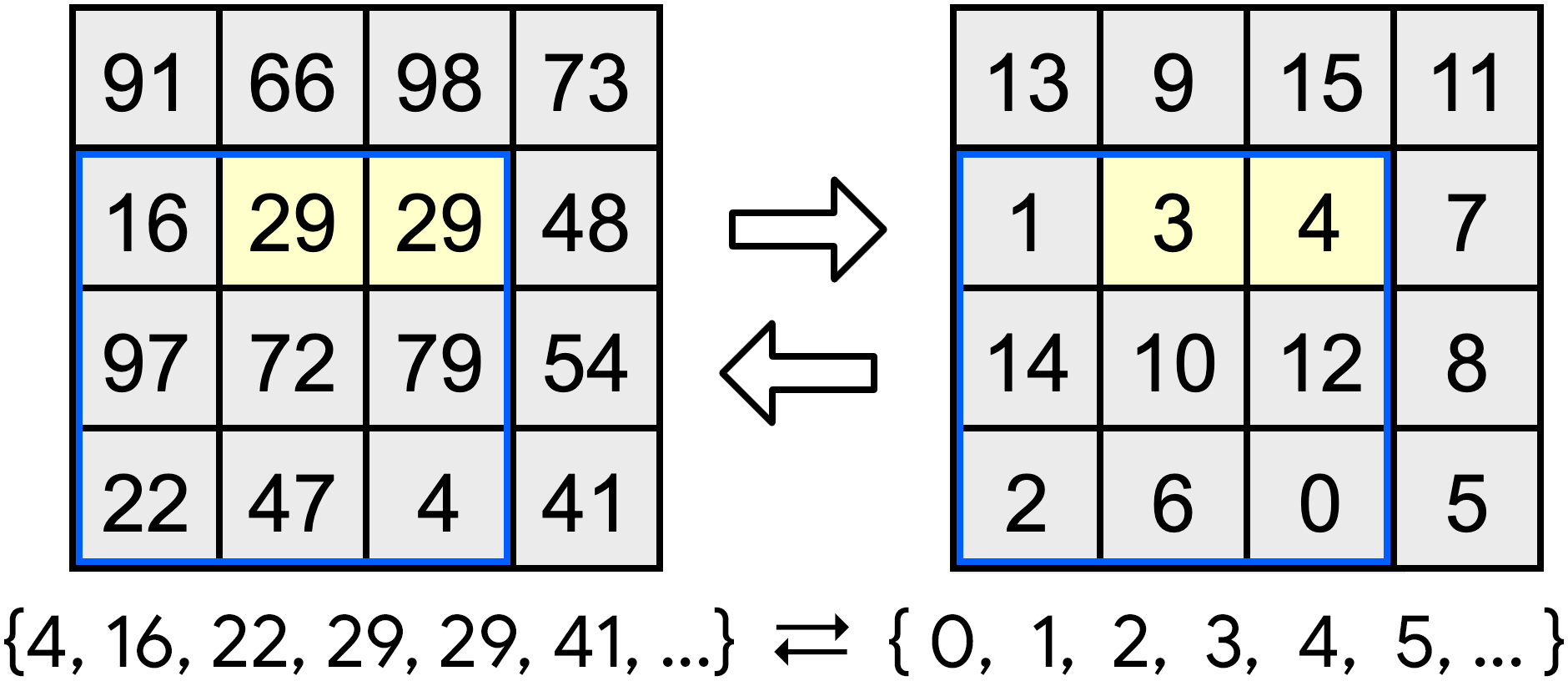}
  \caption{Ordinal transform of Weiss \shortcite{10.1145/1141911.1141918}. Cardinal brightness values (left) map to their respective ordinal ranks (right), creating an "ordinal image." Duplicate cardinal values (yellow) map to consecutive ordinal values. Note that the median values in e.g. the blue squares (29 and 4, respectively) map to each other, illustrating the invariance of the filter under this transform. The lower-left array is the "reverse map" that maps ordinal results back to cardinal output values.}
  \label{fig:ordinal_transform}
  \Description[Illustration of the Ordinal Transform.]{A 4x4 grid of random 2-digit numbers is shown on the left, most tinted light gray, but two numbers that happen to have the same value are tinted yellow. On the right, a 4x4 grid is shown containing the numbers 0 to 15, representing the ranks of the left-side grid values from smallest to largest, with the yellow-tinted squares illustrating that the two matching values \{29, 29\} on the left are mapped to two consecutive ordinal values \{3, 4\} on the right. Below the left "cardinal" grid is the ordered list of the values it contains: \{4, 16, 22, 29, 29, 41, ...\}, and below the right "ordinal" grid is the ordered list of values it contains, which is simply \{0, 1, 2, 3, 4, 5, ...\}.}
\end{figure}

\begin{figure}
  \centering
  \includegraphics[width=\columnwidth]{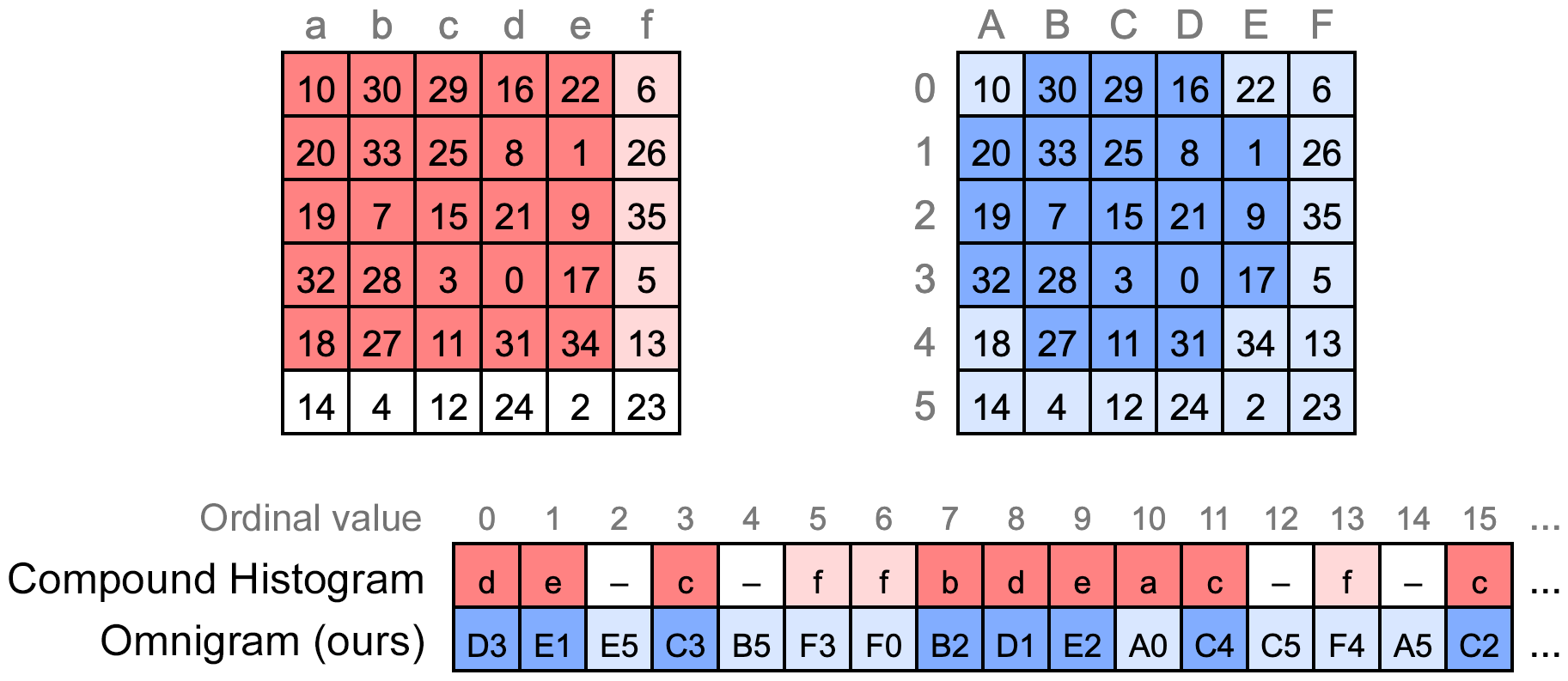}
  \caption{Comparison of the compound histogram (Weiss \shortcite{10.1145/1141911.1141918}) and the omnigram (our method). The compound histogram (red) covers a subset of input rows, encoding the column index of each ordinal value in those rows, or a sentinel '–' for values outside those rows. With the compound histogram, membership of a given ordinal value in an arbitrary full-height rectangle within the red strip can be determined in constant time. With the omnigram (blue), the full 2D location of each ordinal value is encoded, encompassing the entire input image. With this encoding, membership in any arbitrary image region (e.g. the blue circle) can be determined in constant time.}
  \label{fig:compound_histogram_omnigram}
  \Description[Comparison of the compound histogram and omnigram.]{A side-by-side illustration of the compound histogram of Weiss [2006] and our omnigram. Each side contains a 6x6 grid of random 2-digit values; on the left (representing the compound histogram) the top left 5x5 square is tinted red, illustrating that it is a region that can be encoded by the compound histogram, and the top right 1x5 column is tinted light red, illustrating that every such 5x5 square in the top 6x5 block can be encoded by the compound histogram. On the right, inside the 6x6 grid, the top left width-5 circle (as a 5x5 square with the corner pixels removed) is tinted blue, showing that the omnigram can encode the histogram of this circular region. The rest of the entire 6x6 grid is tinted light blue, illustrating that any such circular region within the 6x6 grid can be obtained from the compound histogram.}
\end{figure}

In \cite{10.1145/1141911.1141918} we described an invertible "ordinal transform" between brightness-valued ("cardinal") and rank-ordered ("ordinal") images, as shown in Fig. \ref{fig:ordinal_transform}, replacing duplicate cardinal values with consecutive ordinal values, to yield an "ordinal image" \(I\), that has two particularly useful properties. First, each value in \(I\) is unique, ensuring that histograms of any subregion of it can be represented with just a single bit per element. Second, rank-order operations such as median-filtering are invariant to this transform.

We leveraged these properties to develop a "compound histogram" array, where the value at each index contains the column of that index within the ordinal image, if it is inside the rectangular input strip corresponding to the output scanline currently being processed, or a sentinel value if it is not. From this compound histogram, any binary histogram element of any rectangular subregion within that strip (including every square window) can be obtained in constant time. Once the median value is found for every pixel of the corresponding output row, the compound histogram is slid down one row, with leading values added and trailing values removed, and the next row is processed.

Paired with that paper's hierarchical \(O(log(r))\) method for computing low-precision median filters, the compound histogram approach yields an \(O(log^2(r))\) method for an arbitrary-precision median filter. But it is restricted to square (or rectangular) kernel shapes, and like Huang's method, is limited by the fact that its working set grows linearly with concurrency, because the compound histogram must be updated as the windows are slid vertically. \redline{If several sets of output scanlines are processed concurrently, correspondingly many mutable copies of the compound histogram must be maintained. }This impairs its adaptability to massively-concurrent GPU architectures, where limited fast memory is at a premium.

\subsection{The Omnigram}
Our method extends this "compound histogram" concept to two dimensions. Instead of each array value encoding just a column [or sentinel] value, we encode both the row and column in each element. This doubles the size of the structure relative to the compound histogram, but it has two key benefits. First, it makes the array immutable; any area of the image can be processed with it, not just a single scanline of output, and it never needs to be modified after construction. Second, it allows histogram elements of any image subregion of any shape (not just rectangular shapes) to be extracted in constant time. With these properties, it functions as a universal histogram. We refer to this as our omnigram, \(\Omega_I\).

While processing our image, for any image region \(R\) and any image value \(v\), we want to be able to efficiently obtain the histogram entry:
\begin{align}
H_R[v] &= \text{\# of occurrences of value } v \text{ in region } R
\label{eq:histogram_definition}
\end{align}

Since our ordinal image \(I\) is unique-valued, these histogram elements can only ever be 1 or 0, and our omnigram \(\Omega_I\) specifies where each value is located within \(I\). Thus, this reduces to:

\begin{align}
H_R[v] = 
\begin{cases}
1 & \text{if } {\Omega}_I[v] \in R  \\
0 & \text{otherwise}
\end{cases}
\label{eq:omnigram_point_test}
\end{align}

And for our specific use case of a circular region centered at point \(P\), with radius \(r\), this becomes:

\begin{align}
H_{C(P, r)}[v] = 
\begin{cases}
1 & \text{if } \| {\Omega}_I[v] - P \| \le r  \\
0 & \text{otherwise}
\end{cases}
\label{eq:omnigram_circle_test}
\end{align}

With these foundations, we can distill our method to its essence: we apply a circular modification of Huang's method to the ordinal-transformed image, obtaining histogram values from the immutable omnigram in lieu of maintaining and updating explicit histograms. And instead of sliding a single window at a time, we vertically slide adjacent windows for many columns of output at once. With this approach, each of the algorithm's bottlenecks becomes vectorizable and parallelizable. This enables high performance on both CPU and GPU architectures, while providing the substantial quality improvement associated with isotropic kernels.

\subsection{Method Summary}
At a high level, our method works as follows:

\begin{enumerate}
\item Process the image in small single-plane tiles, with input tiles padded by the filter radius on all sides with respect to the output tiles. (Output tiles may overlap vertically by one row.)
\item For each input tile, perform an ordinal transform [replacing each pixel with its ordinal rank in the tile], also constructing the ordinal->cardinal reverse map and the omnigram \({\Omega}_I\).
\item For a tile at the top of the image, determine the median ordinal value \(m_{00}\) for the top left circular window \(W_{00}\), and store a "pivot" value \(p_0\) near \(m_{00}\), and a "count" \(c_0\) of values smaller than \(p_0\) in this window.
\item Determine the rank of \(p_0\) within the next horizontally adjacent window \(W_{01}\), by comparing leading/trailing values to \(p_0\)  as the window slides one pixel to the right. Then scan \({\Omega}_I\) from \(p_0\) to determine the exact median value \(m_{01}\) for the new window. Select a new pivot \(p_1\)  near \(m_{01}\), and continue sliding horizontally this way to solve the top output row, storing the respective pivots \(p_k\) and counts \(c_k\) for all of the top-row windows.
\item Slide all of the top-row windows down vertically one pixel, in parallel, adjusting each \(c_k\) based on leading/trailing pixels, and using \(p_k\) as the starting point to scan \({\Omega}_I\) for window \(W_{1k}\). Solve the medians for the new row, and update \(p_k\) and \(c_k\) to remain close to the median \(m_{1k}\) for window \(W_{1k}\).
\item Continue this way through all the output rows in the tile. Solutions for the last row can be "forwarded" to the next tile below, since the spatial offset of a solution median pixel from its window center is invariant (for a given global output pixel) between overlapping tiles, and is directly obtainable from \({\Omega}_I\). This allows the relatively expensive steps 3 and 4 to be skipped for subsequent tiles.
\item Invert the ordinal transform using the reverse map to yield the cardinal median-filtered result.
\end{enumerate}

\begin{figure}
  \centering
  \includegraphics[width=0.9\columnwidth]{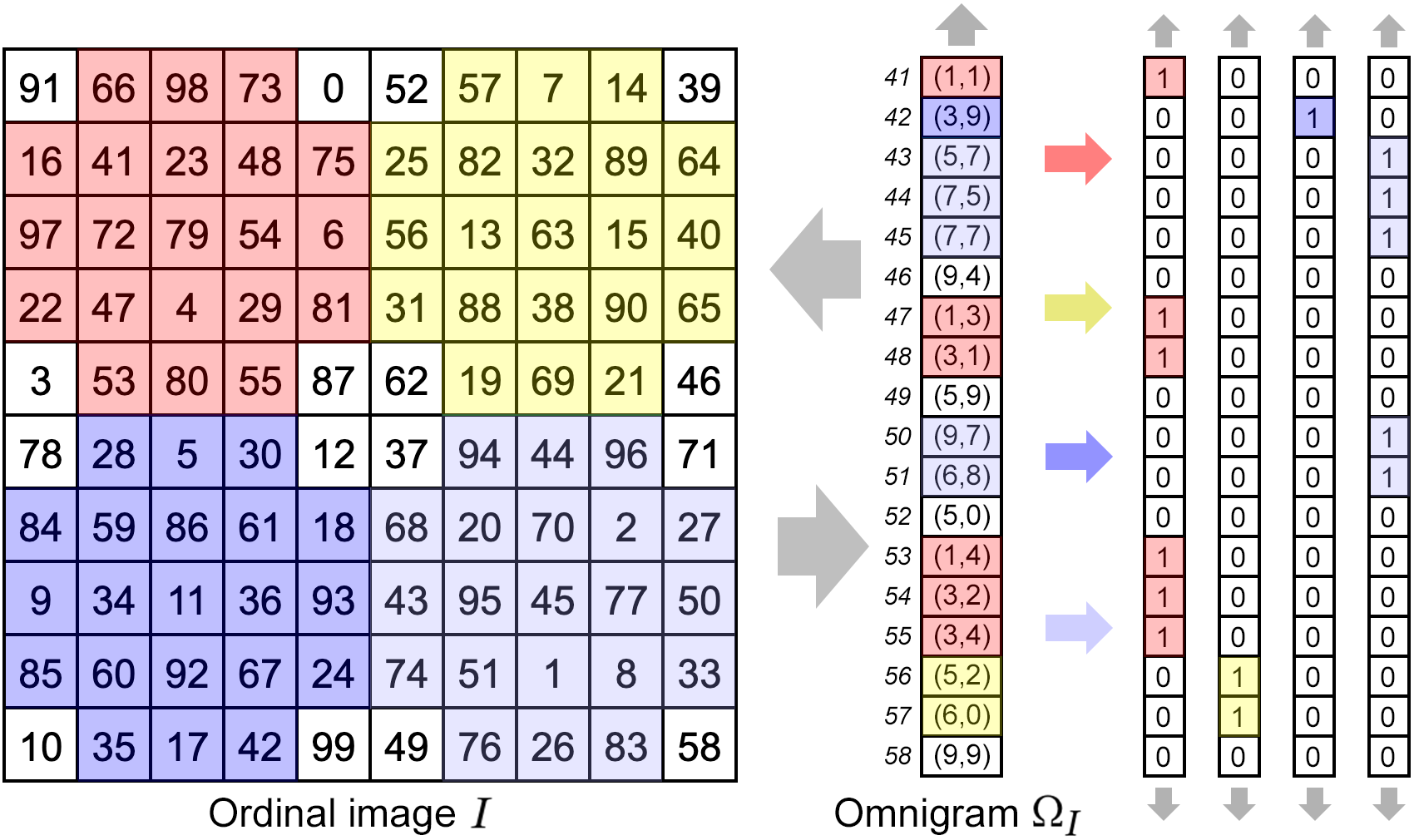}
  \caption{Ordinal image and omnigram. Binary histogram elements (right) of any ordinal image region (tinted circles, left) can be extracted in constant time by querying the omnigram (center).}
  \label{fig:ordinal_image_omnigram}
  \Description[Illustration of the omnigram's functionality.]{On the left, a 10x10 grid of random numbers is shown, with four tinted insets of different colors showing various width-5 circular regions within the grid. In the middle is a column showing the omnigram, which is an array of 2D coordinates; each element of the array is shown tinted corresponding to which (if any) circular region its coordinate belongs to. On the right are four thinner columns showing the binary histograms of the four tinted regions, showing how these binary histograms can be obtained from the omnigram.}
\end{figure}

\begin{figure}
  \centering
  \includegraphics[width=\columnwidth]{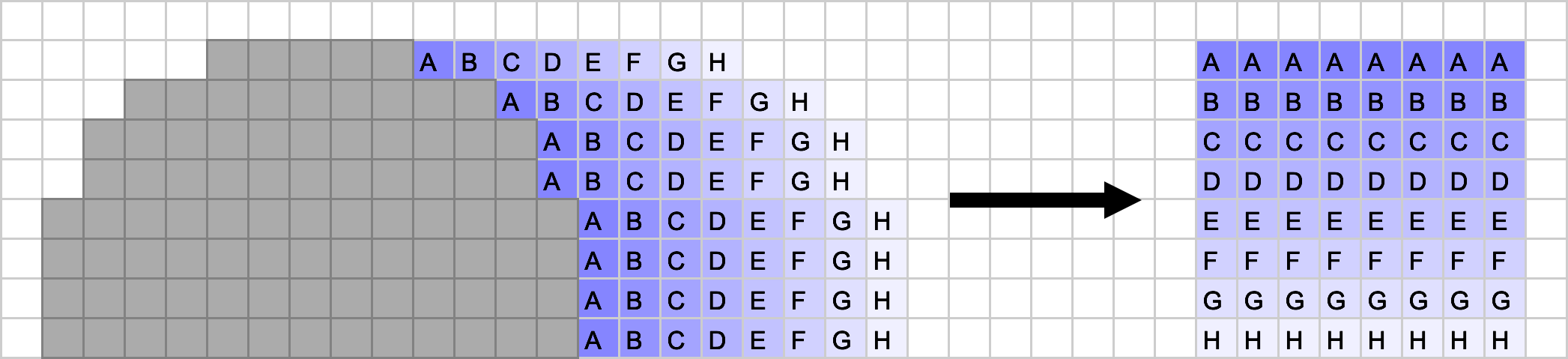}
  \caption{Processing the first row. For sliding the circular kernel window horizontally, it is helpful to shear/transpose the first \(2 * radius + 1\) rows of input pixel data to facilitate SIMD processing. }
  \label{fig:shear_transpose}
  \Description[Visual illustration of a technique for optimizing the first output row.]{On the left, a grid shows the top eight rows of a width-13 circular kernel shape, tinted gray. For each of these eight rows, the pixels immediately to the right of the circular kernel are labeled A, B, C, D, E, F, G, H. On the right of the diagram, a "shear-transpose" illustrates how these rows of values are first sheared into an 8x8 square (this intermediate step is not explicitly shown), then transposed such that all the A's are in the top row, all the B's in the second row, and so forth.}
\end{figure}

\begin{figure}
  \centering
  \includegraphics[height=137pt]{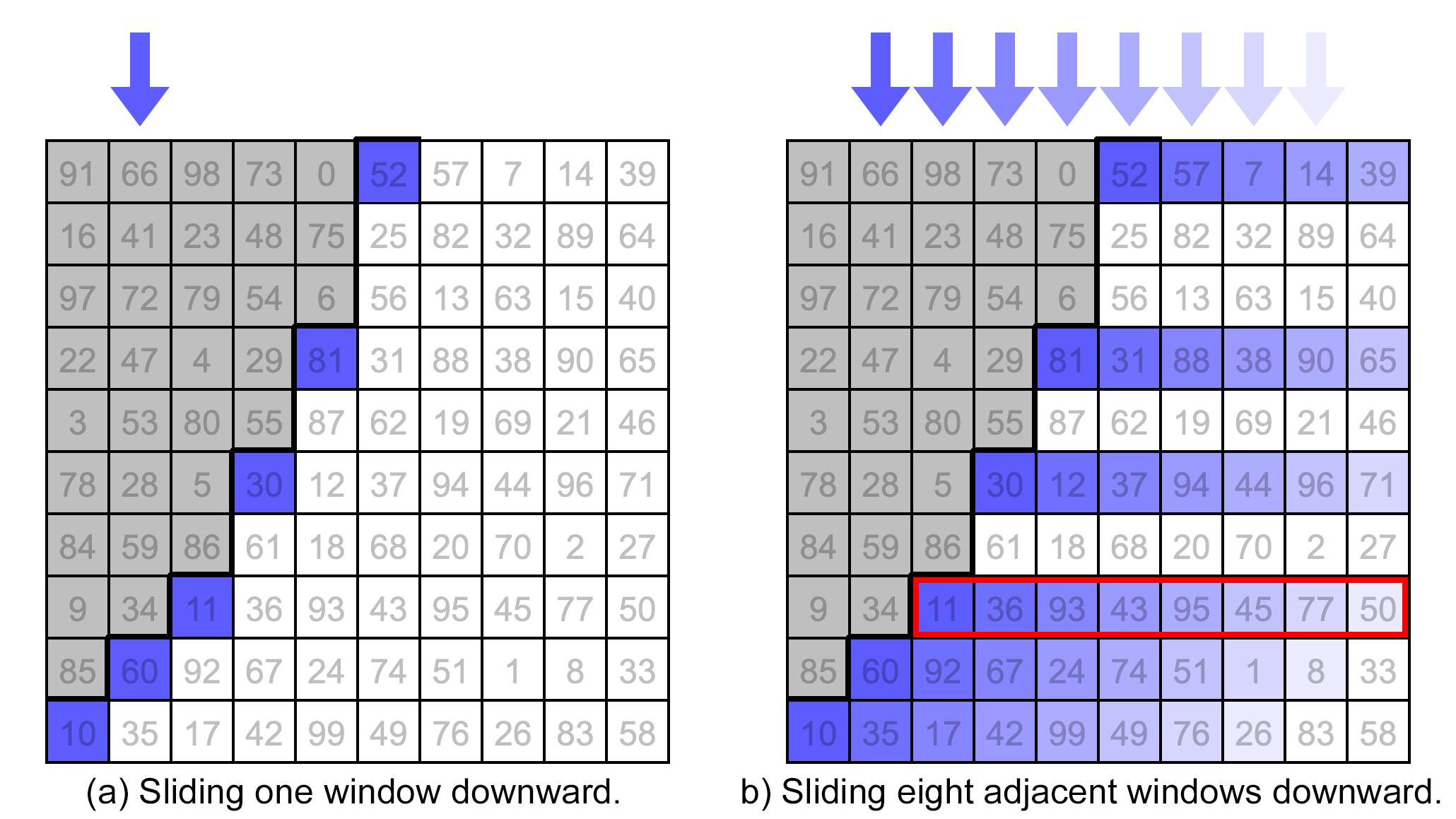}
  \caption{Sliding vertically. (a) When a single non-rectangular window is slid downward, the leading [and trailing] pixels (blue) are scattered in memory, precluding SIMD optimization on a per-window basis. (b) When several adjacent windows are slid downward together, the leading/trailing pixels form coherent groups (e.g. outlined in red), enabling SIMD optimization.}
  \label{fig:slide_vertical}
  \Description[Illustration of how the vertical-sliding step is vectorized.]{On the left, a 9x9 square grid of numbers on the left contains a small "4-o-Clock" section of a large circular kernel in the top left, tinted gray. Within each column, one pixel immediately below the circular kernel shape is tinted blue. On the right, the illustration shows that if eight horizontally-adjacent such circles are processed in parallel, each of the blue pixels on the left side corresponds to an 8x1 section of pixels on the right side. These 8x1 sections can be efficiently vector-processed as the eight windows are simultaneously slid vertically.}
\end{figure}
\section{Implementation}
\label{sec:implementation}

\begin{figure*}
  \centering
  \includegraphics[width=\textwidth]{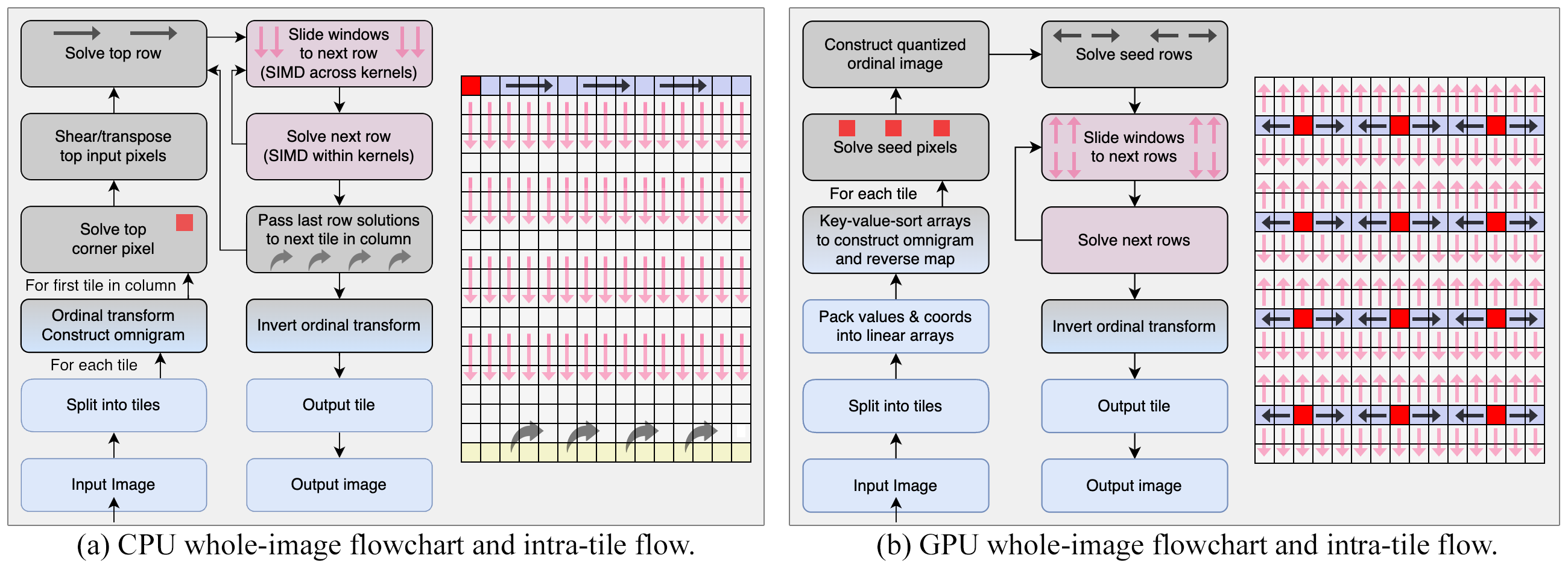}
  \caption{CPU vs. GPU implementations. (a) On CPU, for the top tile in a column, we begin by solving the top left pixel (red), then step across the top row (blue), sweep to the bottom of the tile, then forward the last-row solutions to the next tile. (b) On GPU each tile is processed independently. We solve a sparse grid of "seed" pixels (red), then step left-right to complete sparse rows (blue), then sweep up-down to complete the tile.}
  \label{fig:cpu_gpu_diagram_combined}
  \Description[Flowchart and diagram comparisons of CPU and GPU algorithms.]{On the left half of the figure is the CPU-specific flowchart and diagram; the right half contains the GPU-specific flowchart and diagram. Within each half, the left side shows a 12-step flowchart, and the right half shows an illustration of the pixel grid of the output tile being processed, with the order in which the pixels are processed. This visually summarizes the more complete descriptions given in this section.}
\end{figure*}

We process the image in fairly small tiles, limiting the input tile size to 256x256 pixels. This allows the omnigram to be encoded with 16 bits per element; 8 each for x and y.

Constructing \({\Omega}_I\) and the cardinal-to-ordinal reverse-map for the cardinal input tile \(I_c\) is equivalent to key-value sorting the tuples \{brightness, location\} for each input pixel. The sorted list of cardinal brightnesses \(C\) forms the reverse map; the corresponding list of locations comprises \({\Omega}_I\). If solutions are forwarded between tiles, the key-value sort must be stable to preserve relative coordinate order; otherwise, stability isn't required. Once \({\Omega}_I\) is constructed, the ordinal input image \(I\) is straightforwardly generated as:

\begin{align}
I(\Omega_I[v]) \leftarrow v, \quad \forall v \in \{0, 1, ..., |{\Omega}_I| - 1\}
\label{eq:omnigram_construction}
\end{align}

The vertical window-sliding phase exploits the spatial coherence between adjacent windows shown in Fig. \ref{fig:slide_vertical} to enable efficient SIMD parallelism. Each vector can load several adjacent values from the ordinal image at once, and compare them against the pivots for the corresponding windows, updating the corresponding counts accordingly.

Note that when sliding from window A to window B, Huang's method uses the median value \(m\) of window A as the "pivot" against which to compare entering and exiting values. We observe that the value used for this pivot does not need to be exactly \(m\); it may be any nearby value for which the exact "count" (number of pixels in the window strictly less than the pivot) is known. To facilitate vector alignment while scanning the omnigram, and also because quantized values can be stored in fewer bits, we choose our pivot to be the nearest multiple of 64 to the median value of window A.

After sliding our windows, the subsequent refinement phase scans \({\Omega}_I\) from the pivot index to find the exact solution for a given pixel. To do this, each 16-bit omnigram element must be reduced to a 1-bit histogram element for the relevant window, as per Eq. \ref{eq:omnigram_circle_test} and Fig. \ref{fig:ordinal_image_omnigram}. We perform this reduction in 64-element segments, converting 128 contiguous bytes of \({\Omega}_I\) to a single 64-bit integer bitmask. The  popcount (number of '1' bits) in this bitmask gives us a "coarse" histogram element for the circular region, spanning all 64 indices. If the median value is in this "bin", we scan the bitmask to find the exact ordinal solution \(m\), writing the cardinal solution \(C[m]\) to the output image. If not, we continue scanning \({\Omega}_I\) until the exact solution is found.

As our pivots are always multiples of 64, we observe that if our input tiles are no larger than 128x128, the pivot can be encoded with just 8 bits. We further observe that the low 6 bits of the ordinal image do not affect the result of comparisons with these quantized pivots, so those low bits can also be discarded; we can store just the high 8 bits of the ordinal image as well. For larger input tiles (necessary for filtering kernels above radius \textasciitilde40), we still use 16 bits per ordinal pixel, to preserve the ability to scan \({\Omega}_I\) in 64-element increments.

On CPU, we implement the ordinal transform differently based on the type of the input data. For 8- or 16-bit integer data we use a single-pass bucket sort. For 16-bit inputs, our process for the ordinal transform is as follows:

\definecolor{darkgray}{gray}{0.3}

\begin{itemize}
\item Construct a 65536-element 16-bit histogram $H$ of the tile \(I_c\).
\item Exclusive prefix-sum the histogram: \(H'[k] = \sum_{n=0}^{k-1} H[n]\).
\item Re-scan the tile in row-major order. For each cardinal pixel value \(c\) at \(I_c((x, y)\), select the ordinal value \(v = H'[c]\) and increment \(H'[c]\) by 1. Concurrently, construct the cardinal reverse map \(C\), omnigram \({\Omega}_I\), and ordinal image \(I\):
\end{itemize}

\begin{equation}
\begin{aligned}
  C[v] &\leftarrow p, \\
  \Omega_I[v] &\leftarrow (x, y), \\
  I(x, y) &\leftarrow v \quad \text{\textcolor{darkgray}{// or \(v \gg 6\), in the quantized case.}}
\end{aligned}
\label{eq:ordinal_construction}
\end{equation}

For floating-point images, we implement a modified radix sort to construct the ordinal map. We XOR the exponent and mantissa bits by the sign bit (and flip the sign bit) to allow comparisons of the bit-patterns as unsigned integer, then bucket-sort the top 16 bits in the first pass, followed by bucket-sorting 8 bits at a time to complete the radix sort. In the latter two stages, quicksort is used if only a small number of elements need to be sorted.

\subsection{GPU}

\begin{figure}
  \centering
  \includegraphics[width=\columnwidth]{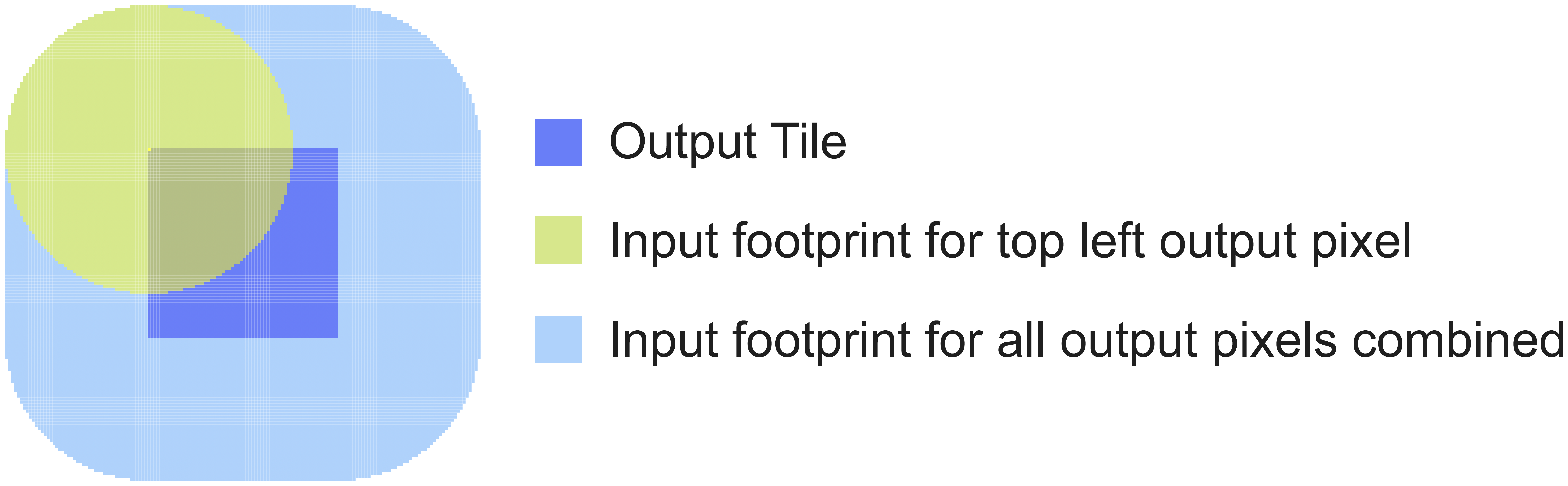}
  \caption{Tile geometry on GPU. For the radius-48 case, each output tile is 64x64 pixels (dark blue). The corresponding input tile footprint (the union of circular kernel windows for every output pixel) is a rounded rectangle (light blue) comprising 23584 pixels, of which 16192 could potentially be the median of any given circular radius-48 window (e.g. yellow). Quantizing pivots to multiples of 64 allows us to cover this range using just 8 bits. (For larger radii, we quantize to multiples of 128.)}
  \label{fig:gpu_48_tile}
  \Description[Input and output tile shapes and kernel shapes, for GPU.]{The diagram shows a dark blue square in the center, representing the 64x64 output tile being processed. Surrounding it is a width-160 light blue rounded rectangle, showing the input tile shape corresponding to this square output region. A radius-48 yellow circle overlay shows the circular kernel window shape corresponding to the top left pixel of the output tile.}
\end{figure}

Our GPU CUDA implementation mirrors the CPU version in several aspects, but with a few key differences. The first of these involves the construction of the omnigram and reverse map. Whereas on CPU we use a custom bucket-sort or radix-sort, on GPU we rely on the highly optimized cub::DeviceSegmentedRadixSort() \cite{CCCL} to sort the entire set of input tiles in parallel. As fast memory is at a premium, we exclude corner pixels that cannot be part of any circular window; each logical input tile is thus a rounded rectangle (as shown in Fig. \ref{fig:gpu_48_tile}, rather than a square. The excluded corner pixels do not participate in the ordinal transform.

CUDA's 32-way warp parallelism informs our choice of tile size; we use output tile sizes of 32x32 for small radii, and 64x64 for larger radii, allocating 256 threads per tile for the former and either 512 or 1024 threads per tile for the latter. As the top-down scanline-by-scanline approach we use on CPU would be insufficient to saturate the GPU's massive concurrency, we instead solve from the middle out, starting from 32 or 64 "seeds" per tile, as illustrated in Fig. \ref{fig:cpu_gpu_diagram_combined}. The median values for the seed windows are solved directly by constructing low-precision histograms from the omnigram (which remains in global memory), scanning these histograms to find approximate median solutions, then refining to exact solutions using the omnigram.

Next, the ordinal image is constructed into shared memory. It is quantized by a 6-bit rightshift for radii <= 48, or a 7-bit rightshift for radii <= 96, enabling it to fit in 8 bits per element. A small amount of additional shared memory is used to store quantized pivots and counts for the seed rows, which are solved by sliding the kernel windows horizontally from the seed windows both left and right; each thread is responsible for one window and one direction. (Some threads are idle in this phase.) When scanning the omnigram to find exact solutions for these windows, the 32 threads in each warp collaborate to scan 64-element blocks of the omnigram for unsolved windows belonging to that warp, until all of the warp's windows have been solved.

Once the seed rows are solved, the seed row windows and new vertical directions (up/down) are reassigned to all threads; these windows are then slid vertically up and down to solve the rest of the tile. With each of the hundreds of threads sharing the same small single immutable ordinal image and omnigram, memory locality is optimal. Finally, as solutions are found, the ordinal->cardinal reverse map is applied to yield the final output image.

\section{Results}
\label{sec:results}

\begin{figure*}
  \centering
  \includegraphics[width=\textwidth]{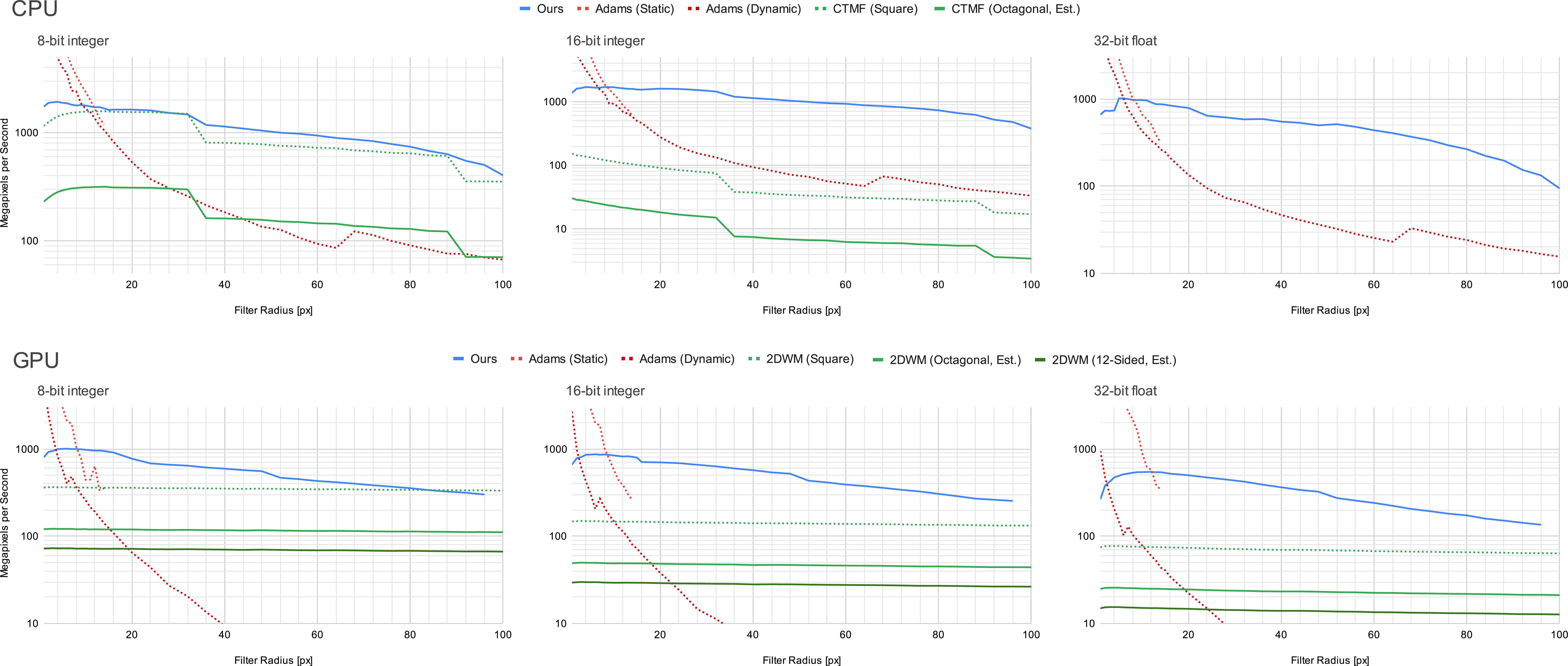}
  \caption{Performance results on CPU and GPU. As our focus is on isotropic filtering, square-kernel implementations are shown with dashed lines.}
  \label{fig:results_cpu}
  \Description[Performance comparison charts for our method versus others.]{Six performance comparison charts are shown: three each for CPU and GPU, each containing performance curves for 8-bit, 16-bit, and floating-point data, respectively, for our method versus previous methods.}
\end{figure*}

We ran our filter's benchmark on a 37-megapixel grayscale image on a 64-core AMD 5995WX CPU with hyperthreading disabled, and on an NVIDIA RTX 4060 GPU. On the same hardware, we ran Moroto's published benchmark code for their 2D Wavelet Matrix filter (2DWM), and also Adams' published benchmark code for their implementation of both their Separable Sorting Networks algorithm and of Perrault's Constant Time Median Filter (CTMF).

 Across all bit-depths and architectures, our method overtakes the fastest SOTA square-kernel implementations between roughly radius 8 and 12, and strongly outperforms SOTA octagonal or 12-sided methods\footnote{Based on Perrault's estimate that their octagonal method is 5x slower than square, and Moroto's estimate that their octagonal and 12-sided methods are 3x and 5x slower than square, respectively.} at all radii, often by over an order of magnitude. On 8-bit images, our method is slightly faster than CTMF's square-kernel implementation on CPU, and outperforms the 2D Wavelet Matrix square-kernel GPU implementation up to radius 84.

 We have also implemented our method for the arm64 architecture, enabling it to run on a wider variety of desktop and mobile devices. And we have benchmarked it on higher-end graphics cards, such as the NVIDIA RTX 5080, confirming that our implementation's performance scales near-linearly in the number of cores. These results are discussed in supplemental material.

Qualitatively, our method provides substantially higher image quality than SOTA square-kernel implementations, as seen in Fig. \ref{fig:teaser} and supplemental images. Fig. \ref{fig:wave_rotated} analyzes this effect quantitatively, showing that the filter's anisotropic artifacts are reduced by well over an order of magnitude with our method, to visually imperceptible levels. Our method may therefore even be preferred over SOTA sorting-network approaches for small kernels (radius < ~8), where the latter retains a nominal performance advantage, at the cost of image quality.
\section{Limitations and Future Work}
Our algorithm's runtime is data-dependent, and thus is theoretically vulnerable to "adversarial" inputs that could result in slow runtimes. (An example would be an input tile that is a high-frequency black-and-white checkerboard, with a thick solid gray perimeter.) We have not observed this with real-world images, but a mechanism to identify and more efficiently process such images might be an area to explore.

A second limitation is that the filter kernel (while isotropic) remains uniformly weighted throughout. Adapting the algorithm to be spatially weighted by e.g. a Gaussian kernel, perhaps along the lines of Kass and Solomon \shortcite{10.1145/1778765.1778837}, might be a useful line for future exploration. Making an "anti-aliased" kernel with smooth falloff at the edges might be another useful improvement.

\subsection{Future Work}
\subsubsection{Coherence}
A key feature of our method is that it relies on the coherence of median-filtered output; the result for each pixel is usually (but not always) similar to the result for the preceding pixel. A systematic statistical analysis of this coherence across a wide variety of images and image types would be useful, to determine optimal parameters for our method, as well as to identify contexts where the image data is more likely to be "adversarial" as described above.
\subsubsection{Small Kernels}
Our main optimization focus has been on medium to large kernels, but we believe there is room for improvement by fine-tuning our method for smaller tiles and smaller kernels. Additionally, we hope this work serves as inspiration for the development of sorting network techniques capable of processing small circular kernels with even higher efficiency.
\subsubsection{Arbitrary Convex Kernel Shapes}
While we anticipate that circular-kernel filtering will be the primary use case for our algorithm, it can be modified to handle a variety of kernel shapes. \blueline{We discuss this further in supplemental material.} \redline{Simple shapes such as polygons (including squares) can be achieved by correspondingly modifying the analytical test in Eq. 3, remaining fully vectorizable on CPU. Or for instance, a "mostly-convex" heart shape could be modeled as the union of two intersecting circles and a triangle. More complex shapes could be achieved by implementing Eq. 2 with a bitmap lookup, which would remain efficient on GPU, but might sacrifice pure vectorizability on CPU, even if most queries could still be solved analytically by testing against e.g. a bounding circle, and only the infrequent exceptions would require a scalar lookup. Shapes of intermediate complexity could be implemented with a "trimap" approach; e.g. computing Eq. 3 against inscribed and circumscribed circles, and applying a bitmap lookup only to points in between those bounds.}
\subsubsection{Ordinal Transform Redundancy}
When the filter radius is a significant fraction of the tile size, \redline{there is another significant area for performance optimization that we haven't yet fully explored, and that is of reducing }\blueline{there are opportunities to reduce} redundancy when ordinal-transforming adjacent overlapping input tiles. E.g., when processing a radius-32 median filter using overlapping 128x128 input tiles (with 64x64 output tiles), each input pixel participates in four ordinal transforms, and thus is sorted four times. Partially sorting the overlapping regions in advance could result in significant speedups for larger kernels, particularly for floating-point images.
\subsubsection{Multi-Parameter Filtering}
Our \redline{current} implementation processes a single radius and percentile value at a time, but it could be straightforwardly modified to compute multiple such values at once, amortizing the cost of the ordinal transform and significantly increasing marginal efficiency. E.g. the filter could return three images with 25\nth, 50\nth, and 75\nth percentile values, \redline{for statistical purposes, or }for variance calculations. Similarly, it could be \redline{easily }modified to process multiple kernel sizes (e.g. radius 8, 16, 32) within a single ordinal transform, mirroring the "runtime-only" efficiency of Moroto et al. \shortcite{10.1145/3550454.3555512}.

\section{Conclusion}
We have introduced a novel fast algorithm for median and general percentile filtering with circular kernels that provides significantly higher image quality and performance than the state of the art, across much of the useful parameter space. By overcoming the limitations of prior methods, we hope our results spur continued research into rank-order-filtering algorithms and their practical and creative applications.

\begin{acks}
We are indebted to Google for providing the resources and environment necessary to pursue this research. Thanks to Yael Pritch, Adi Zicher, Neal Wadhwa, Chloe LeGendre, James Vecore, David Jacobs, Anna Lieb, Frank Barchard, and Artem Belevich for their support and and assistance. Thanks to Michael Herf for his timely hardware wizardry. We also thank the anonymous reviewers for their time and helpful comments. Finally, special thanks to Jenna McGrath for her unwavering support, and for the perfect photo.
\end{acks}

\newpage

\bibliography{references}

\clearpage
\appendix\onecolumn
\begin{figure*}
  \centering
  \begin{minipage}{0.97\textwidth}
  \includegraphics[width=\textwidth]{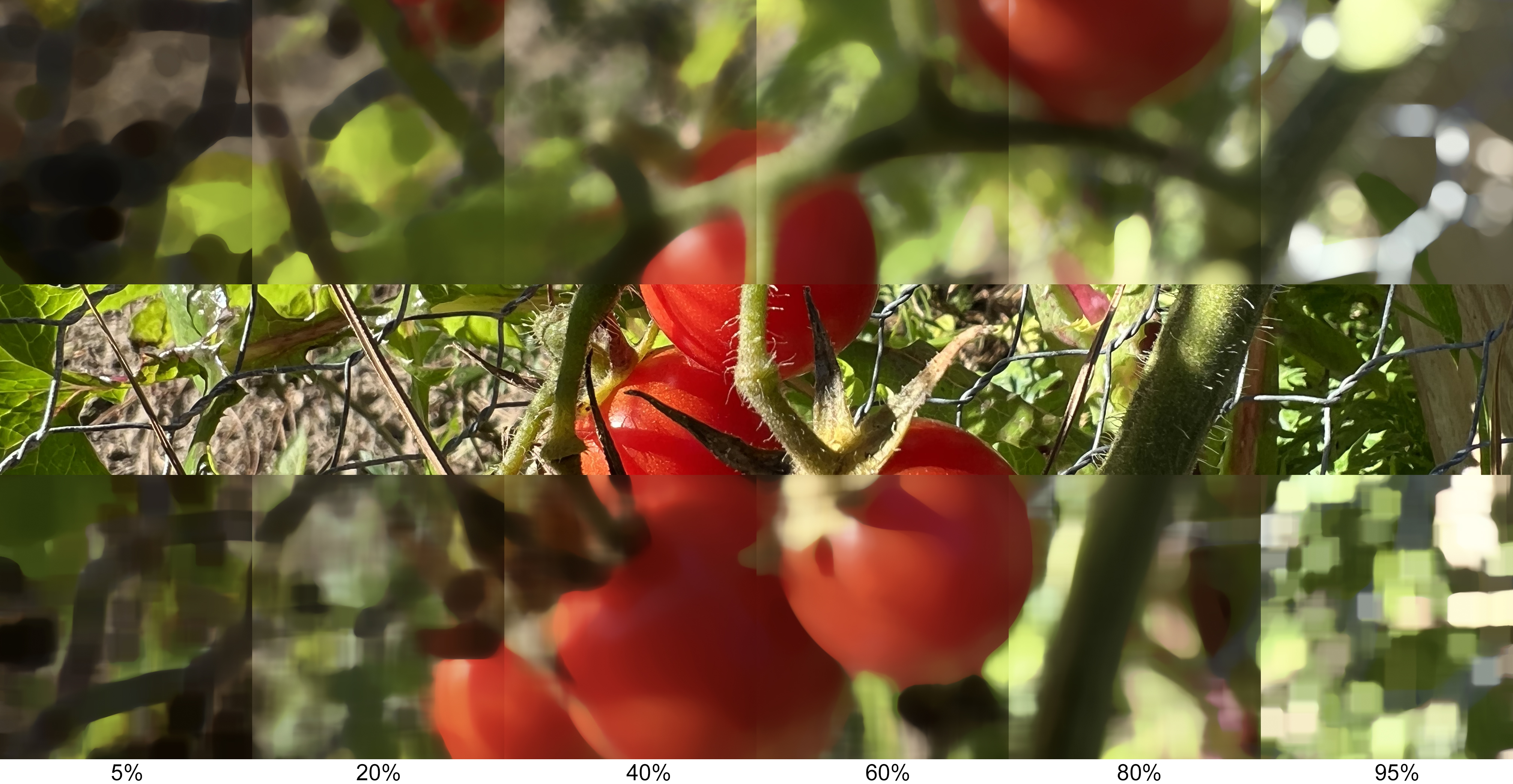}
  \caption{Rank-order "bracketing", for a range of percentiles. Our circular-kernel implementation is shown across the top; an equivalent square-kernel percentile filter across the bottom. The square artifacts are particularly evident toward the endpoints of the percentile range.}
  \label{fig:tomatoes_bracketed}
  \Description[Rank-order "bracketing" visualization on a photo of tomatoes.]{A landscape-orientation image of a cluster of tomatoes on a vine is shown. Across the top and bottom strips, the image is split into six sections, each processed with a different "percentile" filter, from lowest (darkest) on the left to highest (brightest) on the right. The top strip shows the circular-kernel filter; the bottom strip shows the square-kernel filter, and the square-kernel visual artifacts are evident across the bottom strip, but absent in the top strip.}
  \end{minipage}
\end{figure*}

\begin{figure*}
  \centering
  \begin{minipage}{0.97\textwidth}
  \includegraphics[width=\textwidth]{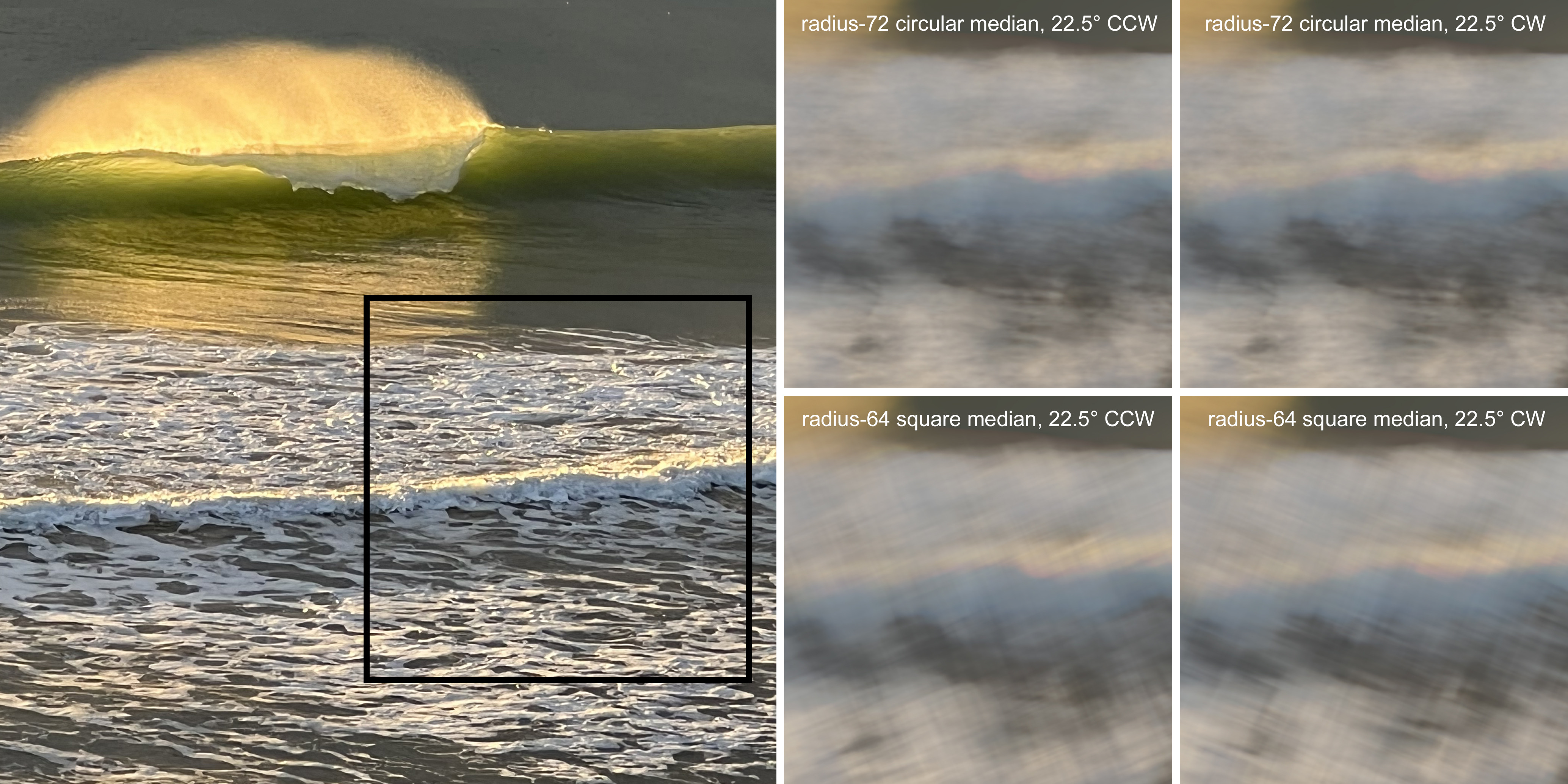}
  \caption{Rotational invariance. The 16-bit input (left) is rotated 22.5° in either direction with bicubic resampling, filtered with equivalent-area square and circular median filters, then rotated back. Our circular-filtered results match each other \textasciitilde35x more precisely under rotation than standard square-filtered results. (Standard deviation 0.06 levels versus 2.12 levels.) The filtered swatches are shown with 200\% unsharp masking to emphasize high frequencies.}
  \label{fig:wave_rotated}
  \Description[Illustration of visual artifacts from processing at different angles.]{On the left is an original photo of a backlit breaking wave, with whitewater in the foreground. On the bottom right are two swatches from the whitewater section, processed by an angled square median filter at different angles, showing how the result is quite different depending on angle. On the top right is the same region processed by the circular kernel at those different angles, showing that the filter is essentially invariant under rotation.}
  \end{minipage}
\end{figure*}

\begin{figure*}
  \centering
  \includegraphics[width=\textwidth]{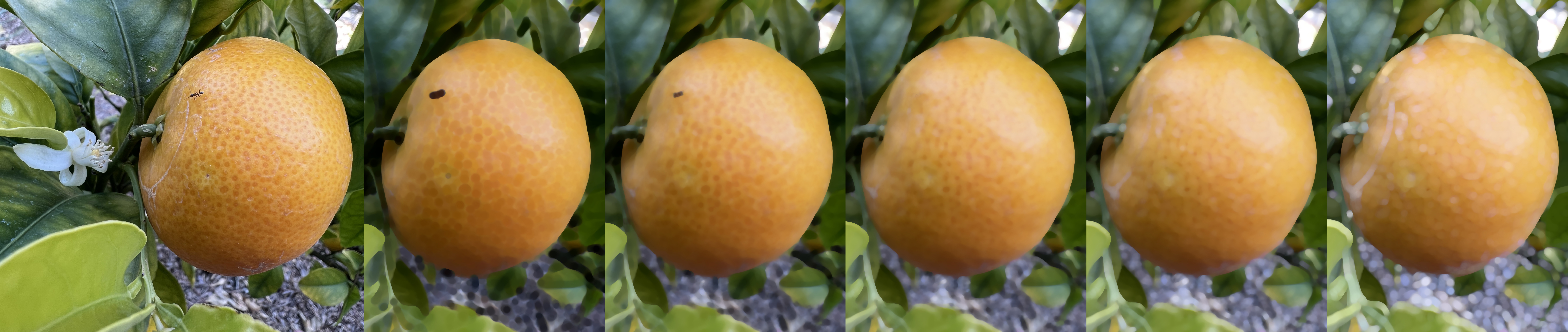}
  \caption{Varying percentiles reveals different image characteristics. Here the input, and 10th, 30th, 50th, 70th, and 90th percentiles, with our radius-48 filter.}
  \label{fig:tangerine_collage}
  \Description[A rank-order "bracketed" set of photos of a tangerine.]{On the left is an original photo of a tangerine. Across the figure, the photo is shown processed with our circular-kernel method at a variety of percentiles, showing how this brings out distinct visual aspects of the underlying photo.}
\end{figure*}

\begin{figure*}
  \centering
  \includegraphics[width=\textwidth]{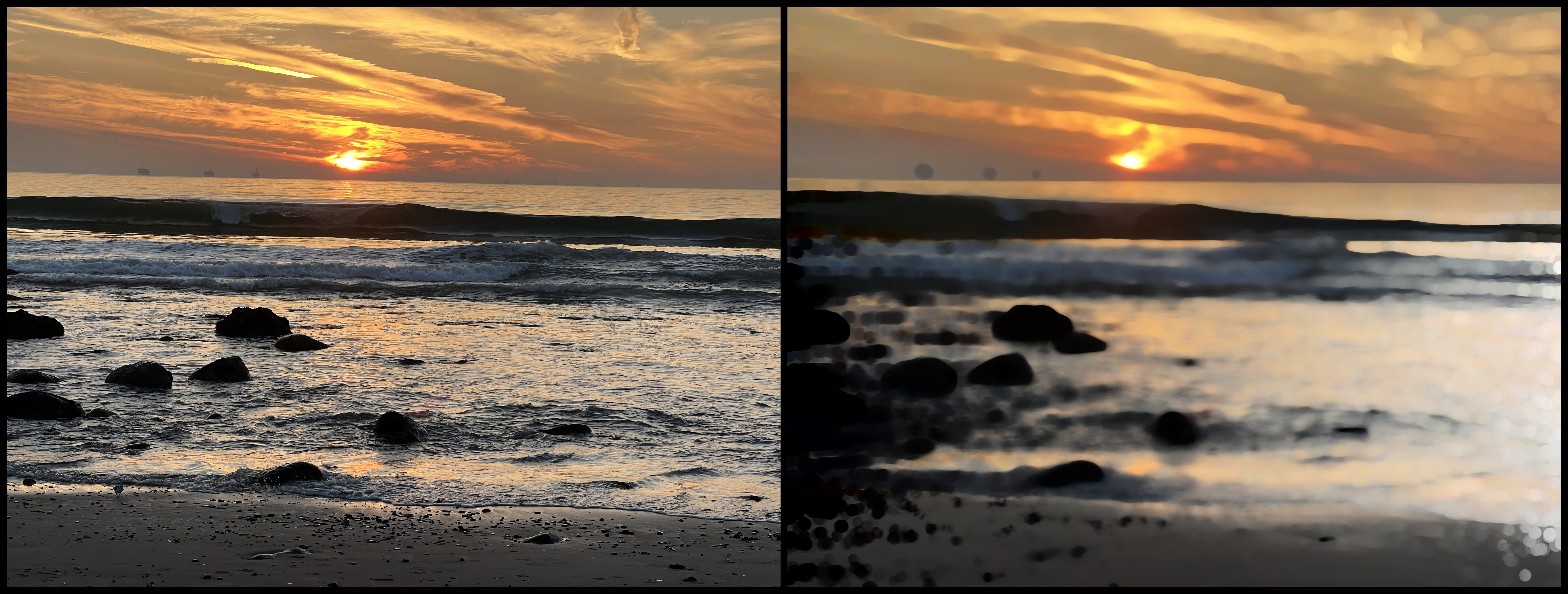}
  \caption{The requested percentile does not need to be constant. Input (left), output (right). Our method can be modified to allow the percentile to vary continuously across the image, shown here varying 0\% to 100\% left-to-right, using our circular radius-32 filter.}
  \label{fig:sunset_collage}
  \Description[Illustration of continuously-varying percentile filtering.]{On the left is an original photo of a beach at sunset. On the right is the same photo processed with our method, with the percentile varying continuously from 0\% (darkest) to 100\% (brightest) horizontally across of the image.}
\end{figure*}

\end{document}